\documentclass[conference]{IEEEtran}
\IEEEoverridecommandlockouts
\usepackage{cite}
\usepackage{amsmath,amssymb,amsfonts}
\usepackage{algorithmic}
\usepackage{graphicx}
\usepackage{textcomp}
\usepackage{xcolor}
\usepackage{caption}
\usepackage{subcaption}
\usepackage{multirow}
\usepackage{hyperref}
\hypersetup{
    colorlinks=true,
    linkcolor=blue,
    filecolor=magenta,      
    urlcolor=cyan,
    pdftitle={Overleaf Example},
    pdfpagemode=FullScreen,
    }
    
\def\BibTeX{{\rm B\kern-.05em{\sc i\kern-.025em b}\kern-.08em
    T\kern-.1667em\lower.7ex\hbox{E}\kern-.125emX}}
\begin{document}

\title{DALLE-URBAN: Capturing the urban design expertise of large text to image transformers \\
}

\author{
    \IEEEauthorblockN{Sachith Seneviratne\IEEEauthorrefmark{1}, Damith Senanayake\IEEEauthorrefmark{2}, Sanka Rasnayaka\IEEEauthorrefmark{3}, Rajith Vidanaarachchi\IEEEauthorrefmark{1}\IEEEauthorrefmark{2}, Jason Thompson\IEEEauthorrefmark{1}}
    \IEEEauthorblockA{\IEEEauthorrefmark{1}Transport, Health, and Urban Design Research Lab,
Melbourne School of Design, University of Melbourne, Australia}
    \IEEEauthorblockA{\IEEEauthorrefmark{2}Department of Mechanical Engineering, Faculty of Engineering and Information Technology, University of Melbourne, Australia}
    \IEEEauthorblockA{\IEEEauthorrefmark{3}Department of Computer Science, National University of Singapore, Singapore
    \\\{sachith.seneviratne, damith.senanayaka, rajith.vidanaarachchi, jason.thompson\}@unimelb.edu.au, \IEEEauthorrefmark{3}sanka@nus.edu.sg} 
}

\maketitle

\begin{abstract}
Automatically converting text descriptions into images using transformer architectures has recently received considerable attention. Such advances have implications for many applied design disciplines across fashion, art, architecture, urban planning, landscape design and the future tools available to such disciplines. However, a detailed analysis capturing the capabilities of such models, specifically with a focus on the built environment, has not been performed to date. In this work, we investigate the capabilities and biases of such text-to-image methods as it applies to the built environment in detail. We use a systematic grammar to generate queries related to the built environment and evaluate resulting generated images. We generate over 12000 different images from 2 text to image models and find that text to image transformers are robust at generating realistic images across different domains for this use-case. 
Generated imagery can be found at the github: \href{https://github.com/sachith500/DALLEURBAN}{https://github.com/sachith500/DALLEURBAN}
\end{abstract}


\section{Introduction}

\begin{figure}
    \centering
    \includegraphics[width=\linewidth]{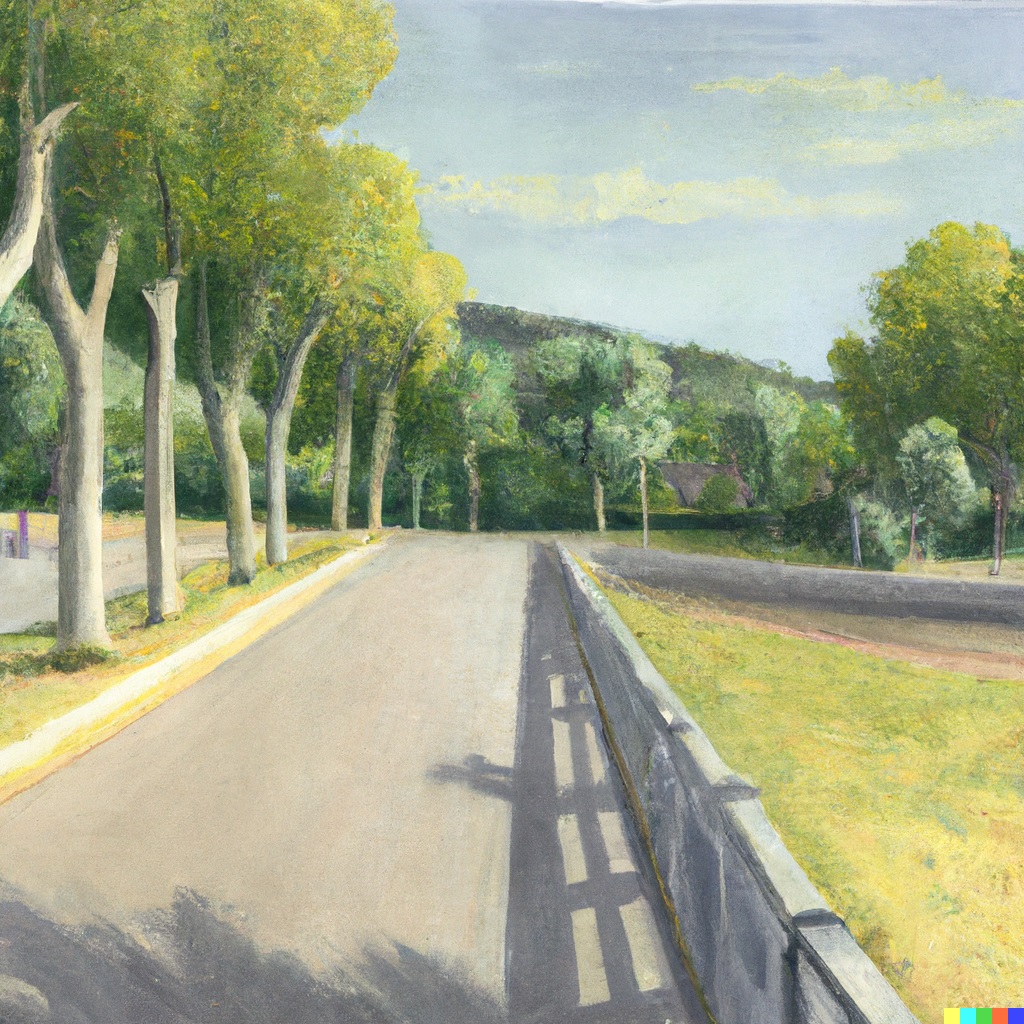}
    \caption{DALL-E 2 generated image for the prompt: 'A painting of a public open space which is ideal for walking'}
    \label{fig:painting_walking}
\end{figure}

The recent advent of deep learning methods that use large text-image datasets, combined using contrastive learning and transformer architectures has led to the possibility of generating realistic images of natural and artificial scenes, such as the image in Fig. ~\ref{fig:painting_walking}. These methods have the potential to aid or automate considerable parts of the visual and built environment design disciplines. However, due to the inherent complexity of such generated scenes, current evaluations of their utility in the literature only capture limited aspects of the potential of such generative methods to disrupt - and potentially aid - applied design disciplines.

An advantage in many neural network-based methods arises in the possibility to reuse the trained networks to instantiate a representation for new tasks. This is frequently exploited in the transfer learning paradigm, which uses a pre-trained model and allows for fine-tuning of the learned weights for a similar (yet potentially more specialized task). For example, ImageNet training is used as the basis for transfer learning for many classification tasks. BERT pre-training is often used as the initialization for many Natural Language Processing (NLP) tasks. In particular, self-supervised learning has been shown to provide many benefits including better label-efficiency, the potential for weaker forms of supervision, and better performance on fine-grained visual tasks (including in the built environment domain\cite{seneviratne2021contrastive}).

However, advances in artificial intelligence frequently have downstream, indirect effects on other disciplines that should be considered. Of these, advances in computer vision, particularly those related to image generation, may have major implications on the visual design disciplines which use drawings, artistic models, prototypes, and even natural scenes. However, previously, design disciplines have been somewhat shielded by a lack of flexibility, creativity and `imagination' demonstrated by AI techniques in combining concepts and where humans have had a clear advantage, these barriers are now possibly beginning to break down. 

Some disciplines create visual artefacts which are more artistic or focused on the aesthetic than the utility - for example art, fashion and jewellery. Alternatively, the built environment disciplines tend to create visual artefacts which can be challenging to immediately evaluate on the basis of quality or utility because their intended impact can span decades or even centuries. Some examples of such analytical creative outputs include precinct structure plans, building facades, designs for public open spaces, and infrastructure layouts that incentivise interaction, healthy behaviour or safety.

In this work we evaluate the potential of large scale, general pre-trained text to image generation models in capturing aspects of the built environment as it relates to health, safety and aesthetics. In particular we evaluate DALL-E 2 \cite{ramesh2022hierarchical} on its ability to handle challenges in urban design and urban planning. We evaluate the model's ability to capture complex concepts such as healthy and safe design, while evaluating biases present in its image generation capabilities. 

Importantly, our work provides an initial evaluation of large pre-trained models' ability to address some of the challenges above \textit{without any domain-specific fine-tuning for the built environment}. Such evaluations are important as it provides a window into future possibilities for fine-tuning with domain-specific and targeted datasets/training. We evaluate the domain coverage of the model over various domains.

Notably, due to the relative infancy of the area of robust text-to-image generation, formal methods to appropriately capture the semantic information present in complex scenes such as built environment scenes do not currently exist. For example, currently proposed evaluation mechanisms \cite{cho2022dall} only take into account certain concepts such as image composition and a limited number of imagery format domains. However, it is important that the potential for such methods to disrupt and/or aid the design disciplines are captured, especially as such methods become more widely available.

In this work, we explore the potential of text-to image generators for investigating urban design qualities by generating a dataset of imagery related to urban design, across multiple formats and geographies. We make our dataset freely available for further evaluation and investigation by the research and urban design community\footnote{\href{https://github.com/sachith500/DALLEURBAN}{https://github.com/sachith500/DALLEURBAN}}.

\section{Background and Related Work}

\begin{figure}
    \centering
    \includegraphics[width=\linewidth]{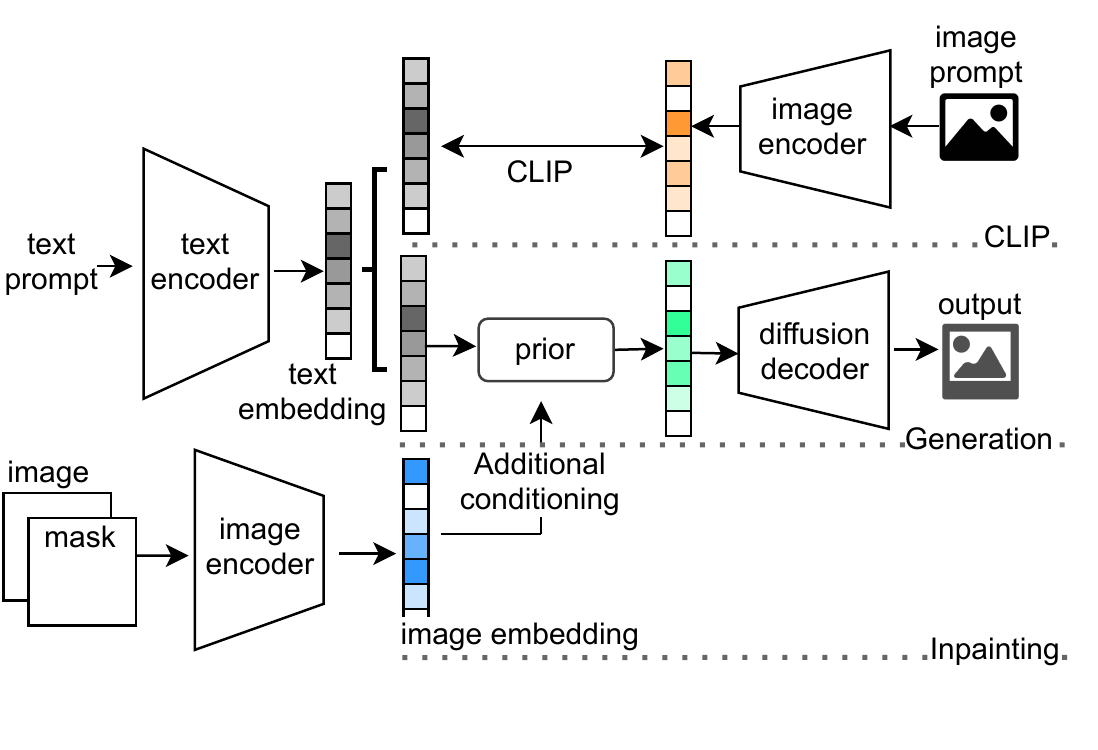}
    \caption{Model architecture. Top: CLIP pretraining, Middle: text to image training and inference, Bottom: inpainting training and inference}
    \label{fig:model}
\end{figure}

\subsection{Artificial Image Generation}

Artificial image generation is a potential means for tackling hitherto poorly understood realms of possibilities in the role of AI in urban design. While the conventional understanding of machine learning as an approach striving for Artificial Intelligence was descriptive (i.e., models that can create an input-to-output mapping based on given data), generative modeling, (i.e., models that can generate data samples) has been proposed as a more robust way to prompt the models to learn the highly relevant representations of the data distributions\cite{esser2021taming}. 

While self-supervised learning methods such as Autoencoders\cite{ng2011sparse} were used for gaining compact representations of data distributions in early literature, modeling the underlying features as probability distributions have been proposed as a generative modeling technique in early attempts such as Restricted Boltzmann Machines (RBM)\cite{larochelle2012learning}. A modern take on this approach, the Variational Autoencoder(VAE)\cite{kingma2013auto} revolutionized the domain of probabilistic modeling of latent feature spaces. However, the conventional take on the VAE does not work well for image modeling due to issues such as mode collapse/posterior collapse, which manifests frequently as blurry backgrounds visible in VAE-generated images\cite{razavi2019generating}. 

A root-cause for this mode collapse is the fact that latent features often follow discrete distributions, not sufficiently captured by the differentiable Gaussian priors. An elegant solution for this was to treat these latent spaces as a mixture of discrete distributions through a quantization of the latent-space. Therefore, vector-quantizing VAEs (VQ-VAEs)\cite{razavi2019generating} were proposed as a superior alternative for generating image samples over conventional VAEs. 

In a more implicit approach for modeling latent-spaces, Generative Adverserial Networks (GANS)\cite{goodfellow2014generative} were proposed as a min-max method to generate data samples from a latent distribution. GANs have two competing subnetworks; a generator that attempts to generate real-like images, and a discriminator that attempts to discern the generated images from actual images. Training these two subnetworks competitively prompts the generator to create images that are difficult for the discriminator to discern as generated. However the modeling of the latent space is indirect, that is, it depends on the discriminator's ability to identify nuanced patterns of features to determine `realness'. Because of this, GANs are later adopted as a supporting technique to increase the realness of VQ-VAE generated results\cite{esser2021taming}. 

More recently, diffusion models have been proposed as a generative modeling technique that identifies a transformation to convert random noise into meaningful images\cite{dhariwal2021diffusion}. This is achieved through two processes, forward diffusion and reverse diffusion. Forward diffusion is a Markov process that adds a small amount of Gaussian noise to an image at each time point. The smallness of the added noise ensures that the noise addition process is reversible, and the Markov assumption helps limit the transition horizon, both of which are required properties to find the reverse diffusion operation. While the forward diffusion process is fixed, the reverse diffusion process is parametrised and approximated in a small-step fashion (i.e. starting from complete random noise, a reduction of a small amount of noise is conducted at each step until a meaningful image emerges). The images generated through diffusion processes are found to be closer to what humans would perceive to be real\cite{nichol2021glide}. 

In addition to DALL-E 2, recent methods such as `Imagen'\cite{saharia2022photorealistic} and `Parti' \cite{yu2022scaling} by Google and `Make a Scene' by Meta \cite{gafni2022make} are also notable text-to-image models. More recently, Stable Diffusion (SD) using Latent Diffusion Models \cite{Rombach_2022_CVPR} has been proposed as an open-source counterpart of DALL-E 2. Although no quantitative comparison using metrics such as the Frechet Inception Score is done, SD is considered to generate text-to-image outputs with comparable performance. Imagen uses a diffusion model similar to GLIDE and both Parti and Make a Scene uses a VQ based image reconstruction method similar to VQGAN. Notably, both Make a Scene and Imagen address a different shortcoming present in GLIDE and DALL-E 2. Imagen is capable of generating better contextual images suitable to the captions \cite{saharia2022photorealistic} and Make a Scene is capable of generating `out-of-sample' images adhering to the caption, although the caption may violate the perceived reality through the training set \cite{gafni2022make}. Parti, on the other hand uses a sequence-to-sequence architecture based fully on transformers to generate `image tokens' from `text-tokens', and reconstruction is carried out from these image tokens by using a similar strategy to the VQGAN method. This is an alternative to the diffusion upsampling at the early stage, which allows the direct creation of mid-resolution images ($256 \times 256$) as opposed to the $64\times64$ images that the diffusion decoders generate in DALL-E 2. Importantly, the potential for fine-tuning the text embedding of such models\cite{gal2022image} or end-to-end fine-tuning of the entire model while minimizing textual drift\cite{ruiz2022dreambooth} is of considerable interest for customized urban design and for increasing, reducing or removing the bias within the model.

While we focus on text to image generation, the advent of text to 3D diffusion models via either Neural Radiance Fields\cite{poole2022dreamfusion} or 3D objects\cite{sanghi2021clip} and text to video diffusion models\cite{https://doi.org/10.48550/arxiv.2209.14792} create promising new directions for design oriented research and applications. Diffusion models for generating graphs\cite{https://doi.org/10.48550/arxiv.2209.14734} also have considerable potential in this regard (for example road network synthesis).

\begin{table}
    \centering
    \caption{Inpainting example 
    \label{tab:picturegrid}}
        \begin{tabular}{cc}
            Original & Mask \\
            \includegraphics[width=0.4\linewidth]{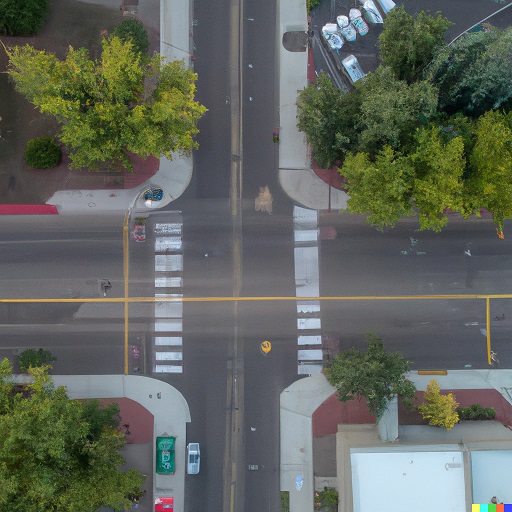} & \includegraphics[width=0.4\linewidth]{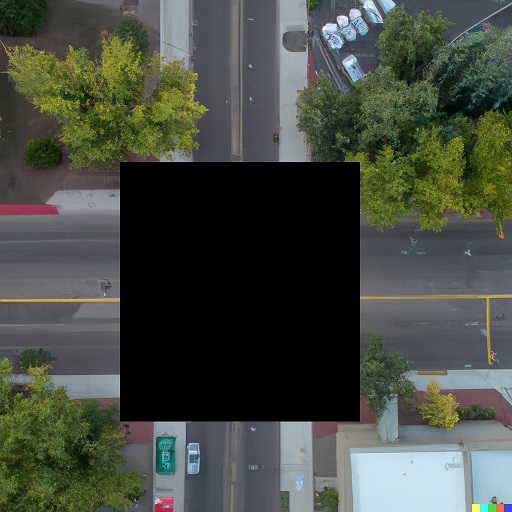} \\
            \multicolumn{2}{c}{In-painted with: ``A safe intersection"} \\
            \includegraphics[width=0.4\linewidth]{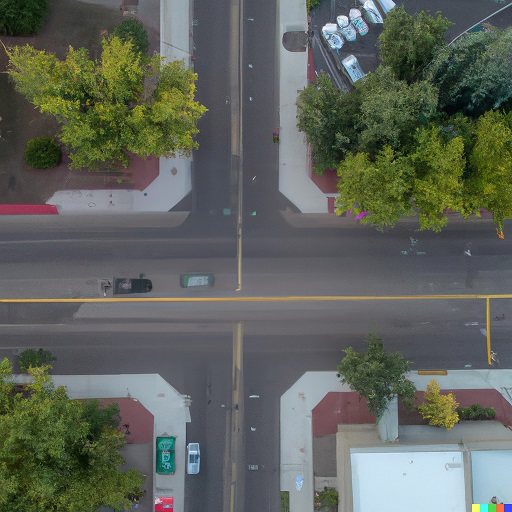} & 
            \includegraphics[width=0.4\linewidth]{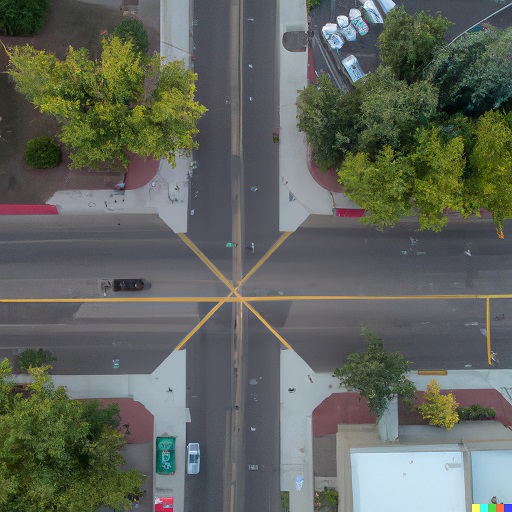} \\
            \includegraphics[width=0.4\linewidth]{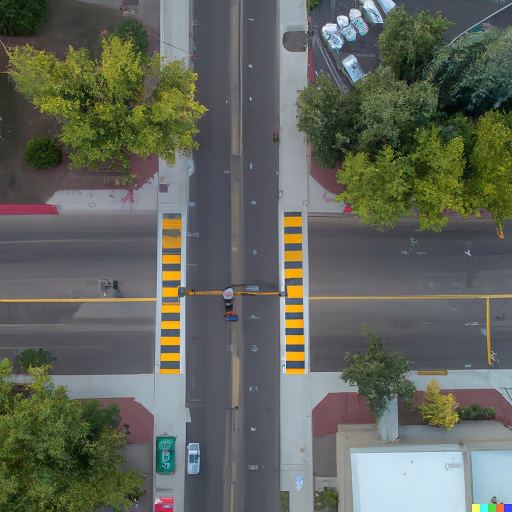} &
            \includegraphics[width=0.4\linewidth]{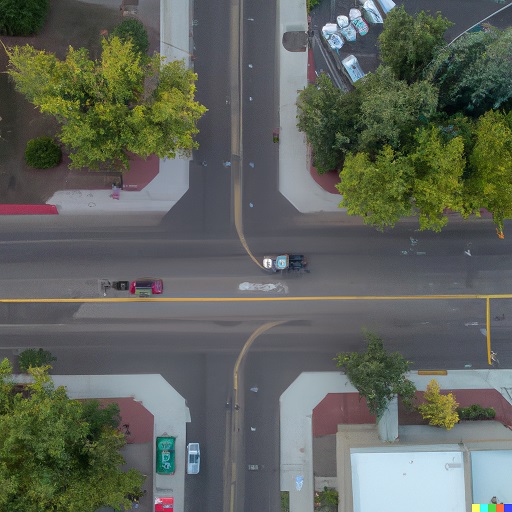} \\ 
            \multicolumn{2}{c}{\includegraphics[width=0.4\linewidth]{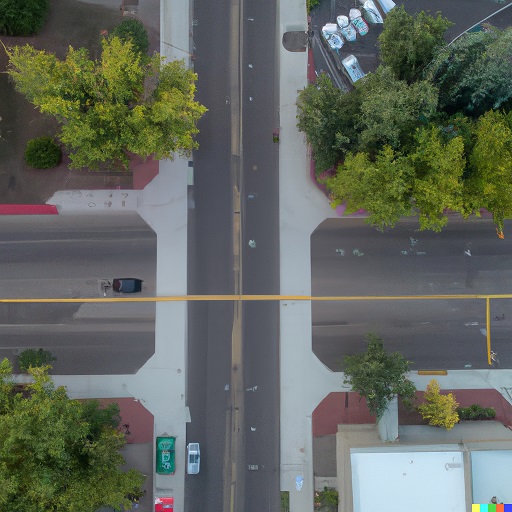}}\\
        \end{tabular}%
\end{table}

\begin{table*}[t]
    \centering
    \caption{A sample of the grammar rules used for generating queries. By varying the terms within square brackets, it is possible to compare generative behavior. For each comparison, only the square bracket(s) of a single rule is allowed modification\label{tab:grammar}}
        \begin{tabular}{|p{0.09\linewidth} | p{0.3\linewidth} | p{0.2\linewidth} |p{0.4\linewidth}|}
        \hline
        Query type & \hfil Purpose & \hfil Example of grammar rule & \hfil Examples  \\
        \hline
        \multirow{3}{\linewidth}{Format(\textbf{F})}
           &  \hfil \multirow{3}{\linewidth}{\hfil Captures the ability of the model to generate realistic imagery in different format domains}
           & \hfil F $\rightarrow$ G & `\textbf{(empty string)}a neighborhood people would enjoy walking in' \\ 
           \cline{3-4}
           & &\hfil  F $\rightarrow$ `A [Format] of ' + G & `\textbf{a painting of }a neighborhood people would enjoy walking in' \\ 
           \cline{3-4}
           & & \hfil  F $\rightarrow$ G + `, in [Format]' & `a neighborhood people would enjoy walking in\textbf{, in sketchup}' \\
 \hline
 
 \hfil \multirow{3}{\linewidth}{Geography(\textbf{G})}
           &  \hfil \multirow{3}{\linewidth}{\hfil Captures the ability of the model to generate realistic imagery to represent different geographic locations and styles}
           & \hfil G $\rightarrow$ Q & `a healthy neighborhood\textbf{(empty string)}' \\ 
           \cline{3-4}
           & &\hfil  G $\rightarrow$ Q + ` , [City],[Country]'  & `a healthy neighborhood\textbf{, Sydney, Australia}' \\ 
           \cline{3-4}
           & & \hfil  G $\rightarrow$ Q + ` in [Continent]' & `a healthy neighborhood\textbf{ in Antarctica}' \\
 \hline
 \hfil \multirow{3}{\linewidth}{Urban Quality(\textbf{Q})}
           &  \hfil \multirow{3}{\linewidth}{\hfil Generating scenes containing desirable or undesirable urban characteristics}
           & \hfil Q $\rightarrow$ U & `a \textbf{(empty string)}neighborhood' \\ 
           \cline{3-4}
           & &\hfil  Q $\rightarrow$ `[Adjective] ' + Q & `a \textbf{healthy safe} neighborhood' \\ 
           \cline{3-4}
           & &\hfil  Q $\rightarrow$ Q + `with [Property]' + & `a city\textbf{ with a high level of pollution}' \\ 
 \hline
 
 \hfil \multirow{3}{\linewidth}{Urban Unit(\textbf{U})}
           &  \hfil \multirow{3}{\linewidth}{\hfil Captures variation in the urban unit of analysis}
           & \hfil U $\rightarrow$ [`city'] & `a \textbf{city} with a high level of pollution' \\ 
           \cline{3-4}
           & &\hfil  U $\rightarrow$ [`park'] & `a \textbf{park} which is ideal for children'  \\ 
           \cline{3-4}
           & & \hfil  Q $\rightarrow$ [`neighborhood'] & `a safe \textbf{neighborhood}' \\
 \hline
        \end{tabular}%
\end{table*}

\subsection{Evaluation of generated images}
The advances in image synthesis methods have led to higher quality results. However, most of these results are corroborated by qualitative evidence. When it comes to quantifying the quality of generated images objectively and comparing between different approaches, it is essential to have a proper framework for evaluating quality.

The existing metrics to evaluate generated images often rely on a network pretrained on large real datasets. For example, popular classic metrics such as \textit{Inception Score (IS)} and\textit{ Fréchet Inception Distance (ID)} utilise the InceptionNet architecture pretrained on ImageNet. 

In IS, the outputs of the InceptionNet model when the generated images are provided as input are used to calculate Kullback–Leibler divergence of the conditional and marginal class distribution. Despite its popularity, it has been shown that this metric is sensitive to modal parameters and implementation \cite{odena2019open} and can be biased towards ImageNet class labels \cite{barratt2018note}.

FID calculates the Fréchet distance between the multivariate Gaussians of the embedding space in the Inception-v3 model. FID gained popularity due to its consistency with human inspection. However, its been shown that FID does not perform well with smaller sample sizes \cite{chong2020effectively}. 

We forgo the assessment of the generated images through these metrics as neither IS nor FID are able to take into account the perceived `realness' and the quality of the images.

New variations of these classic measures are proposed with an attempt to address their shortcomings. Spatial FID (sFID) \cite{nash2021generating}, Class-Aware FID (CAFID) \cite{liu2018improved}, Memorization-informed FID (MiFID) \cite{bai2021training} and Unbiased FID/SI \cite{chong2020effectively} are some such metrics. 

Some qualitative metrics moving away from traditional methods propose having a human in the loop for the evaluation. Human Eye Perceptual Evaluation (HYPE) \cite{zhou2019hype} and Neuroscore \cite{wang2020use} approaches aim to have an evaluation that is closer to human inspection. The advancement of generative models has led to the advancement of these evaluation metrics which provide a better method to compare between models \cite{borji2022pros}. However, none of these metrics allow us to consider the caption relevance, ability to seamlessly edit/inpaint with given inputs and compare images from format domains such as cartoons.

With the increased volume and quality of images being produced by DALL-E, there are now some DALL-E specific evaluations emerging. For example, Cho et. al. \cite{cho2022dall} propose Dall-Eval which incorporates visual reasoning skills, text alignment and quality, and social biases. This shows that as generative models become more complex and capable, the evaluation metrics also should improve to capture these.

\subsection{Design of the built environment}

Acknowledged characteristics of the built environment including configuration, aesthetics, and diversity directly affect human experience, perceptions of health, and objectively measured health outcomes. For example,  urban greenery \cite{Herrmann-Lunecke2021} contributes to notions of “health” and “happiness” associated with the environment. Other metrics such as “safety”, “wealth”, “walking-friendliness” are also associated with urban environments that demonstrate given characteristics. Hence a positive human perception of the environment is a critical goal to be satisfied in the design process for built environments. \cite{Gu2021} identify the importance of the individual creative cognitive process, alongside the human cultural aspect and collaboration (3Cs) as contributing factors to the nature of environments and built form. They also identify three ways in which computers are involved in the creative process of built environmental design: Computational creativity (CC), Human-Computer Co-Creativity (HC3), and Creativity Support Tools (CSTs). While potentially perceived as threats by some \cite{Novak2018}, we recognise the potential of DALL-E 2 like applications in contributing to the HC3 and CST-type computational design and therefore, the generation of environmental and urban design concepts that - if implemented - may enhance the health and well-being of populations.

\subsection{Computer Vision methods and their utility in the built environment}

The exploration of image generation in built environment design has so far been limited, with the only work exploring the conversion of urban areas based on characteristics related to health\cite{wijnands2019streetscape}. More focus is instead present in capturing (primarily using classification, detection and segmentation approaches) elements of the built form for commercial, inventory or database applications, or on occasion relating to the development of city and neighborhood typologies\cite{THOMPSON2020e32}.

Contrastive Representation learning (which forms the basis for CLIP) has been used in prior work to explore aspects of the built environment. Advances in this regard as related to remote sensing imagery\cite{manas2021seasonal} often has implications on analysing the built environment. Identifying species distributions based on environmental characteristics has been attempted using contrastive learning\cite{seneviratne2021contrastive}. The ability to geolocate an image to the city of its origin based on urban style is an important problem in urban computing and has been explored using contrastive learning\cite{seneviratne2021self}.

\begin{table*}[h]
    \centering
    \caption{Domain coverage.\label{tab:domains}}
        \begin{tabular}{cccc}
        \multicolumn{2}{c}{Image types} & \multicolumn{2}{c}{Geographic Locations}  \\
             \includegraphics[width=0.25\linewidth]{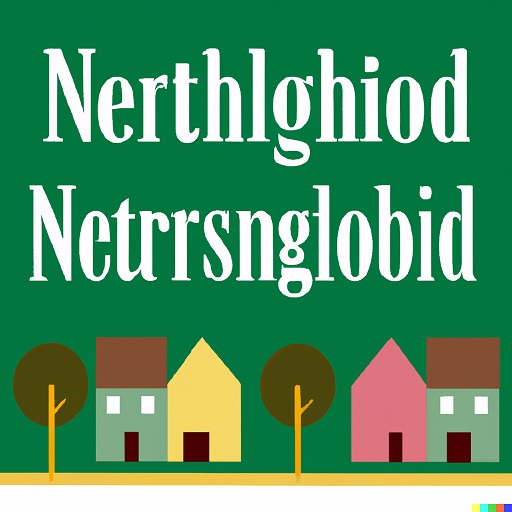}
            & \includegraphics[width=0.25\linewidth]{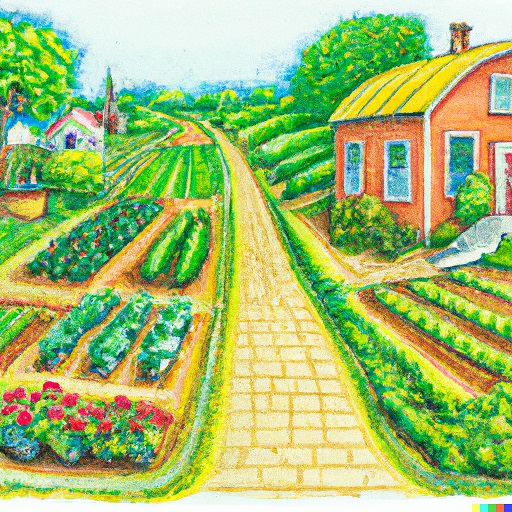}
           & \includegraphics[width=0.25\linewidth]{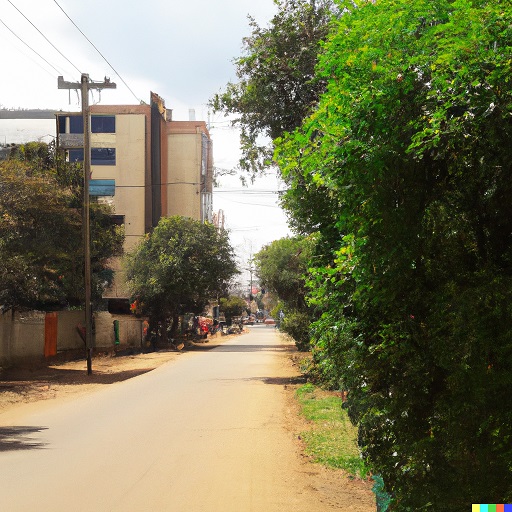}
            & \includegraphics[width=0.25\linewidth]{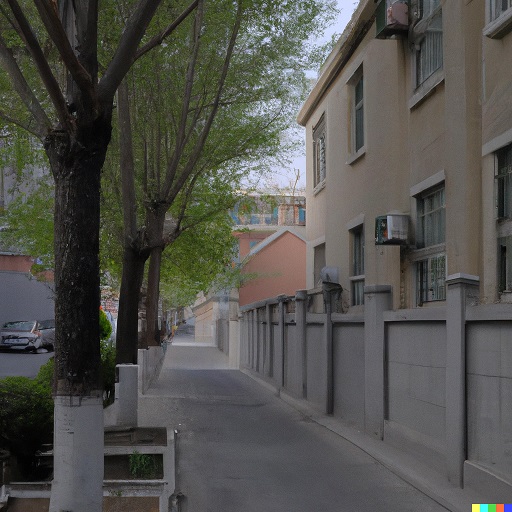}\\
            (a) poster & (b) painting & (e) Nairobi, Kenya  & (f) Beijing, China \\
              \includegraphics[width=0.25\linewidth]{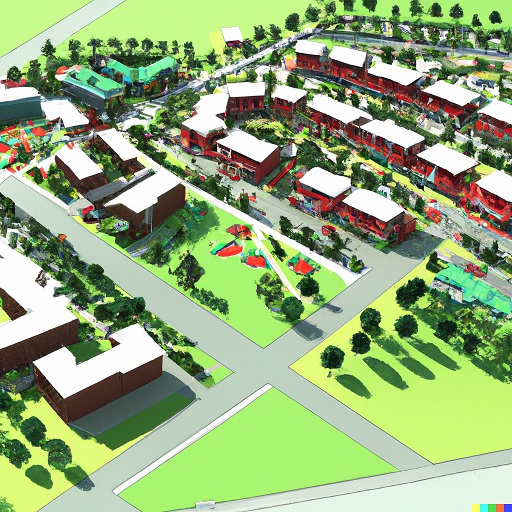}
            & \includegraphics[width=0.25\linewidth]{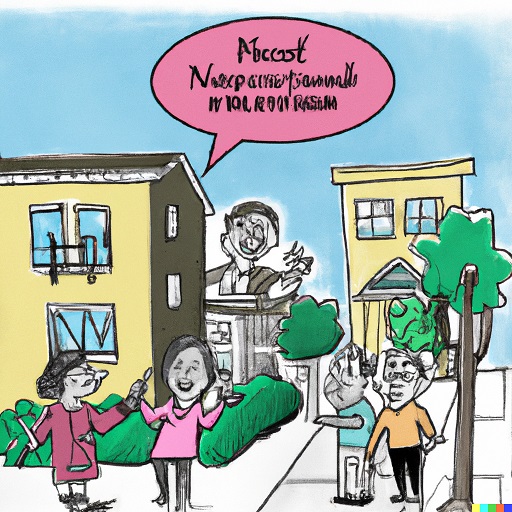}
            & \includegraphics[width=0.25\linewidth]{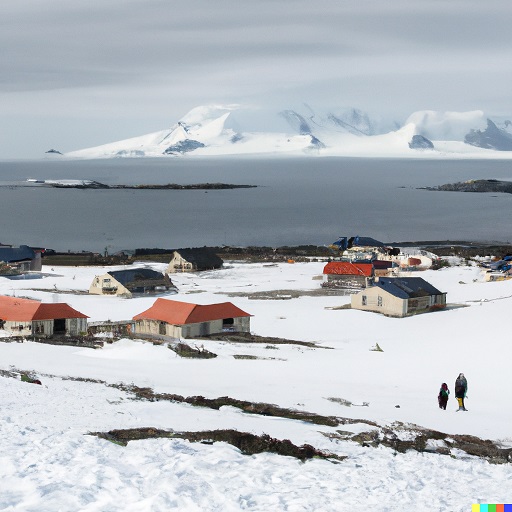}
            & \includegraphics[width=0.25\linewidth]{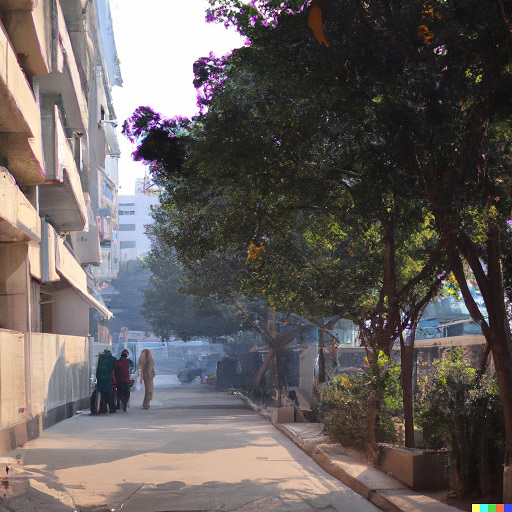} \\
            (c) architectural model & (d) cartoon & (g) Antarctica & (h) Delhi, India \\
        \end{tabular}%
\end{table*}

\section{Methodology}

\subsection{Model architecture}

To generate images from a text-prompt, the model requires three main components; 1) A model that encodes text prompts as possible captions into a latent embedding (text encoder), 2) a model that maps caption text embeddings to a latent space of image embeddings (prior) and 3) a model that generates images from the image embeddings (image decoder). Both DALL-E and DALL-E 2 frameworks contain these three components. In the original DALL-E framework, the first and second components of the  model were implemented as a Contrastive Language and Image Pre-Training (CLIP)\cite{radford2021learning} model. The CLIP model is trained on a paired dataset of images and captions, and the model is trained to predict the image given a caption and vice-versa by passing the caption through a language encoder and then passing the images through an image encoder. With a pretrained CLIP model, a text-to-image-embedding workflow was completed, and for the generation of images from the image embeddings, the decoder component of a VQ-VAE was used. In the DALL-E 2 workflow, the image embeddings were decoded using a diffusion decoder as used in GLIDE, which is a closer predecessor to DALL-E 2 than the original DALL-E. The key differences in DALL-E 2 and GLIDE lie in their training is that the DALL-E 2 leverages an additional CLIP objective to better guide the text-encoding training in addition to the GLIDE training process. 

The first step of training the DALL-E 2 model is training its CLIP model to obtain a good text-to-image correlation in the text-encodings (i.e., training the text encoder). This is done  using the 400M image/caption pairs from the CLIP training set and 250M image/caption pairs from the DALL-E training data. Once the CLIP model is trained, the CLIP encoder is frozen, and a prior (text-encoding to image-encoding model), decoder (image-encoding to $64 \times 64$ image) and an upsampler ($64 \times 64$ image to $1024 \times 1024$ is trained using the latter 250M images due to the noisiness present in the former 400M.) During this training the text-encoder in the CLIP model, a transformer with a causal attention mask is used which has 24 transformer blocks.

There are two methods considered for training the prior; 1) an autoregressive prior (AR) and 2) a diffusion prior. The diffusion prior was observed to outperform the autoregressive prior on aspects of photorealism, caption similarity, and diversity of images. To include picture-editing capabilities to the trained model, the authors employ a fine-tuning specifically catered to inpainting tasks. During this fine tuning, random regions from the training images are cropped-out, and the remainder is produced to the model with an additional channel for the region annotation mask. To further improve the upsampling process of the inpainting model, while the full image of the low-resolution is provided when training the upsampler, only the unmasked region of the high-resolution image is provided. Fig.~\ref{fig:model} highlights the training and inference process.

The diffusion decoder was a modified GLIDE model with 3.5B parameters. In addition to the text-condition pathway of the original GLIDE model, the CLIP encodings were also concatenated to the GLIDE text encodings to allow the model to learn aspects from both text-encoding methods. Because of the use of the CLIP embeddings to generate images outside of reranking images based on caption similarity, DALL-E 2 is considered to be an unCLIP modification of GLIDE, rather than a direct successor of DALL-E version 1.0. 

In our work, we generate images in a systematic manner using two separate generation techniques (text-based generation and text-based in-painting). 

\subsection{Inference - image generation from text}

As per the pipeline, during the image generation from text task, the text captions are passed through the text-encoder, which outputs a CLIP text-embedding. This embedding is then passed through the prior which generates an image embedding, which is then used to condition a diffusion decoder to output an image. 

For text-based generations, the top 6 filtered images matching the text prompt are selected. The grammar for generating queries is further discussed in Section~\ref{sec:grammar}. We used the following observations and associated reasoning for designing the grammars for the selected queries:

\subsection{Inference - completing an image based on a text prompt}
The model facilitates completing an image based on a text prompt among other image manipulations. In addition to the CLIP embeddings generated from the text-to-image workflow, in the completion task, the model takes in additional 4 channels of input (RGB channels and a mask channel). During the inference, the encodings of the masked input image are then concatenated to the CLIP embeddings and fed to the prior. In this training formulation, the model learns to fill in the masked parts of the image. This `inpainting' workflow was used primarily to validate the observations made through the previous worklow (image generation conditioned only on the text prompt).

For inpainting, the top 5 filtered images are selected and evaluated by comparing to the original image used for the generation and the text prompt used in the inpainting process. An illustrative example can be found in Table~\ref{tab:picturegridinp}. The inpainting process has led to the modification of the original image by addition of pedestrian crossings, more signal lights while maintaining the overall coherence of the image. In particular, this workflow has potential for correcting any errors introduced in the text-conditioned generation process, using a human-in-the-loop workflow.


\subsection{Experimental Design}
\label{sec:grammar}
We follow existing evaluations of DALL-E\cite{mishkin2022risks} in performing an evaluation of the quality of the generated images. We use a simple rule based system for evaluating different aspects of the generated images. A subset of the rules is given in table~\ref{tab:grammar} for illustrative purposes. We explore 4 separate related dimensions of urban design: the image format domain(F - for example `painting'), the geography(G - for example `Melbourne, Australia') of generated images, the urban quality(Q - for example `healthy') and finally the urban unit of analysis(U - for example `neighborhood'). Creating a query starts from the Domain(D) and involves expanding the terms in the order F $\rightarrow$ G $\rightarrow$ Q $\rightarrow$ U. This ordering is followed because the urban unit generally forms the subject of the query, with immediate adjectives/qualities considerably changing the interpretation of the query. The Geography(G) has less impact than immediate urban qualities, but needs to be incorporated into the query prior to constructing the image format domain. For example, 'A poster of ((Q), in Nairobi, Kenya)' is a sensible way of formatting the query Q with parentheses indicating the order of evaluation.

\section{Results and Evaluation}

For evaluating the performance of the model, we treat the task performed by the model as an information retrieval (IR) problem. This is motivated from an applied design perspective, where a single relevant image from the prompt can serve as the starting point for further design or modification by either a human or a system with a human-in-the-loop. Therefore, when the model returned at least one image which was deemed to be relevant to the query, we treat this as a success. Out of the 170 queries performed, only 2 queries returned images which were completely unrelated/misinterpreted.

Importantly, the model is able to capture variation in the format domain as well as geographical area with considerable realism, as shown in Table~\ref{tab:domains}(a-d) generated using a query of the form 'a [FORMAT] of a healthy neighborhood'. The ability to capture geographic style is portrayed in Table~\ref{tab:domains}(e-g) which was generated using 'a neighborhood people would enjoy walking in, [CITY], [COUNTRY]'.

Generated imagery often captured well-known aspects in urban design: for example, the relationship between green spaces and health as shown in Table~\ref{tab:picturegrid}. The absence of green space in the antonym query('unhealthy') was helpful in verifying this observation. This was further verified by inpainting results as shown in table~\ref{tab:picturegridinp}(a $\rightarrow$ b-d). A similar observation was that 'safe' neighborhoods were associated with higher quality buildings (fresher paint, completed construction), while unsafe areas were associated with the opposite. This was also verified via inpainting (table~\ref{tab:picturegridinp}(e $\rightarrow$ f-g).
\begin{table}
    \centering
    \caption{Healthy neighborhoods (top) vs unhealthy (bottom)\label{tab:picturegrid}}
        \begin{tabular}{ccc}
             \includegraphics[width=0.3\linewidth]{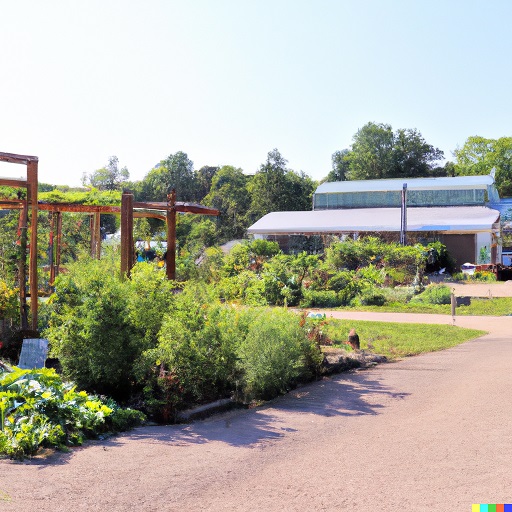}
            & \includegraphics[width=0.3\linewidth]{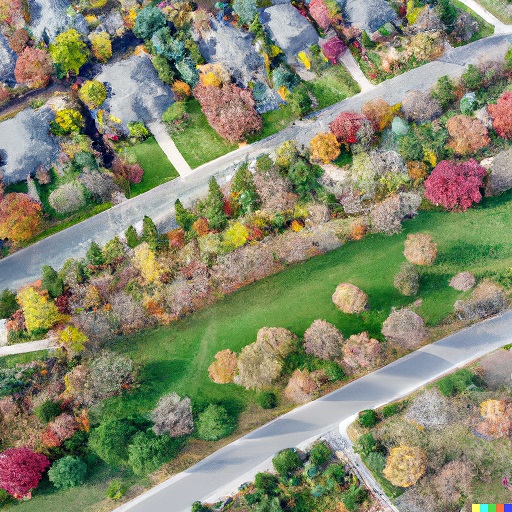}
            & \includegraphics[width=0.3\linewidth]{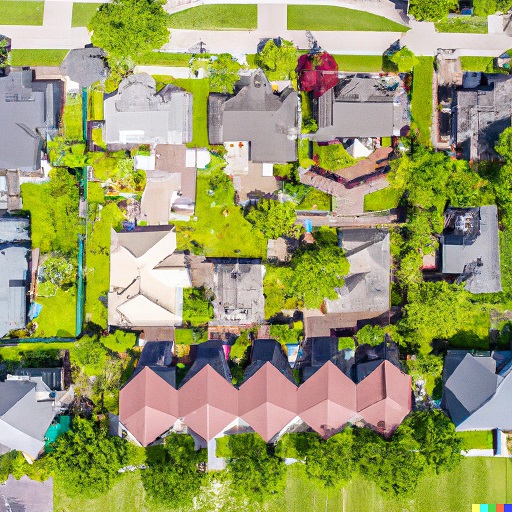}\\
            (a) & (b) & (c) \\
             \includegraphics[width=0.3\linewidth]{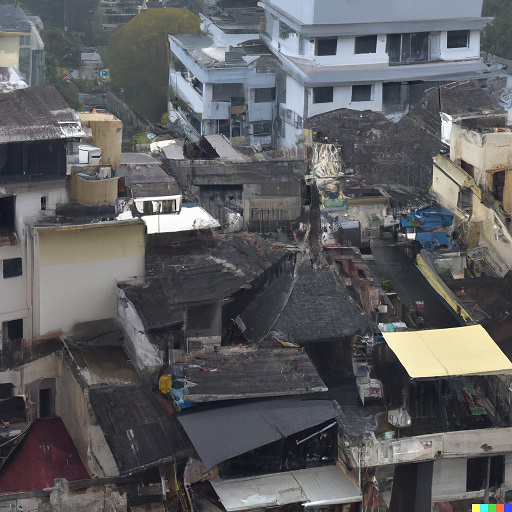}
            & \includegraphics[width=0.3\linewidth]{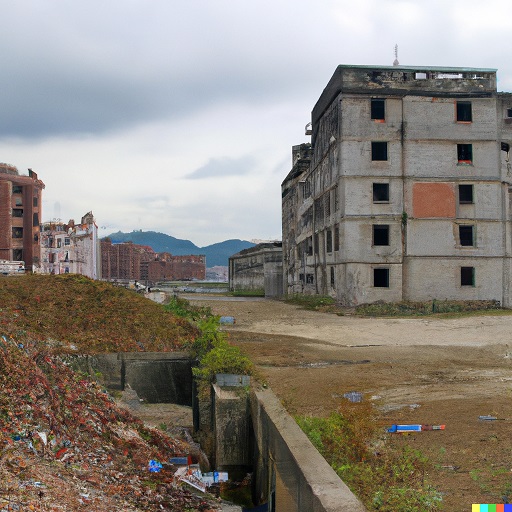}
            & \includegraphics[width=0.3\linewidth]{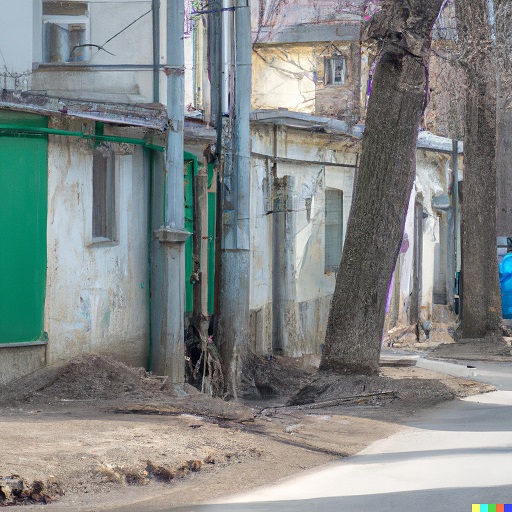}\\
            (d) & (e) & (f) \\[-2pt]
        \end{tabular}%
\end{table}

\begin{table*}[h!]
    \centering
    \caption{In-painting as a means of validating the findings from text conditioned generation.\label{tab:picturegridinp}}
        \begin{tabular}{cccc}
        \multicolumn{2}{c}{Healthy (a) $\rightarrow$ Unhealthy (b-d) (trees inpainted)} & \multicolumn{2}{c}{Unsafe (e) $\rightarrow$ Safe (f-h) (upper half inpainted)}  \\
             \includegraphics[width=0.25\linewidth]{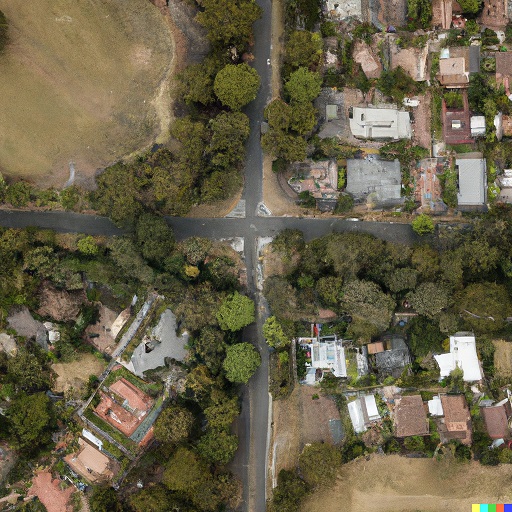}
            & \includegraphics[width=0.25\linewidth]{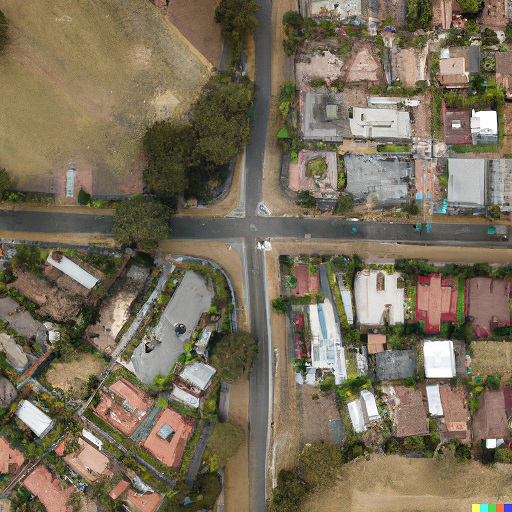}
            & \includegraphics[width=0.25\linewidth]{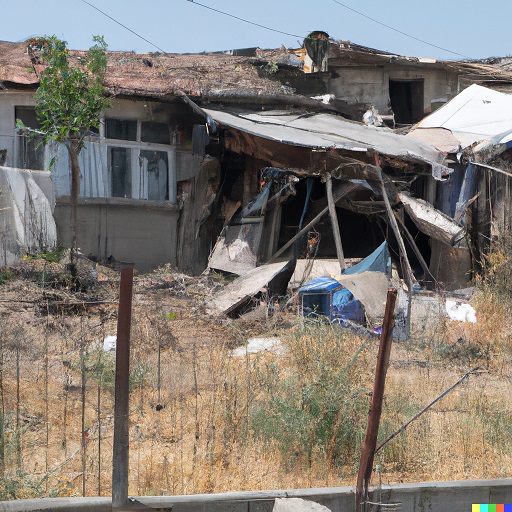}
            & \includegraphics[width=0.25\linewidth]{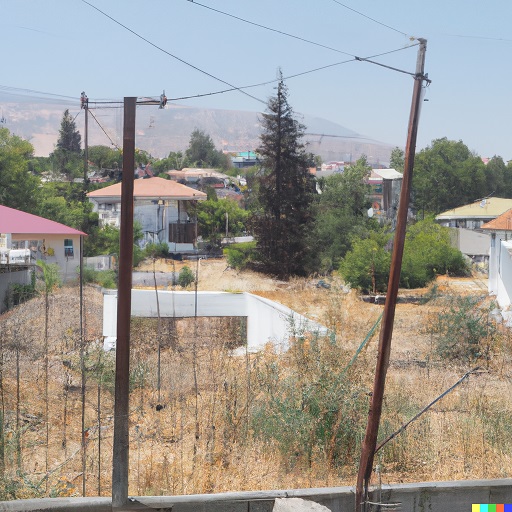} \\
            (a) & (b) & (e) & (f) \\
            \includegraphics[width=0.25\linewidth]{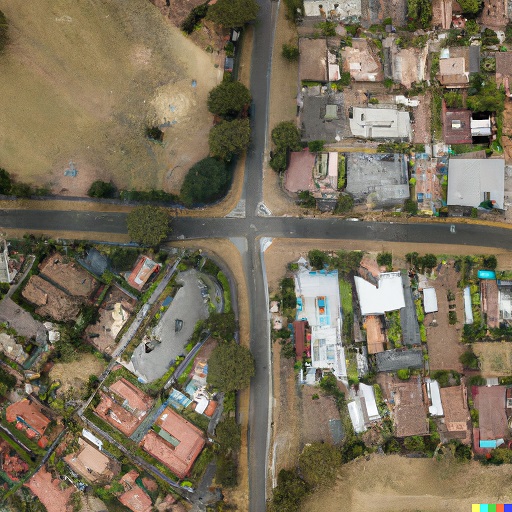}
            & \includegraphics[width=0.25\linewidth]{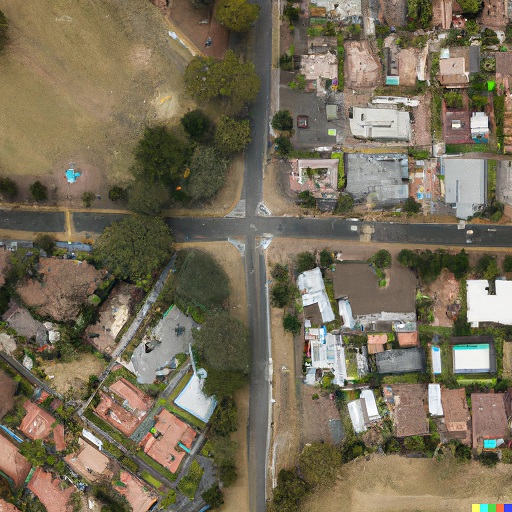}
            & \includegraphics[width=0.25\linewidth]{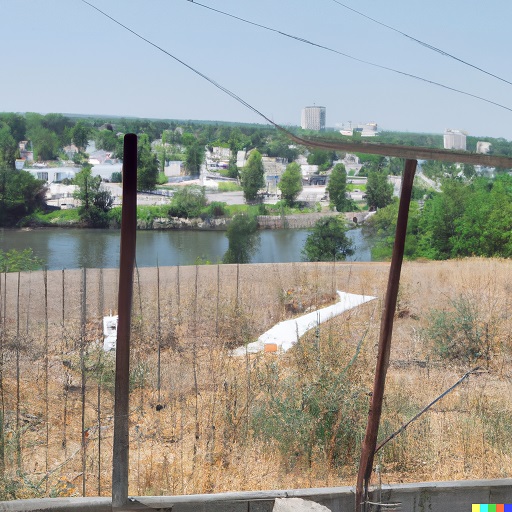}
            & \includegraphics[width=0.25\linewidth]{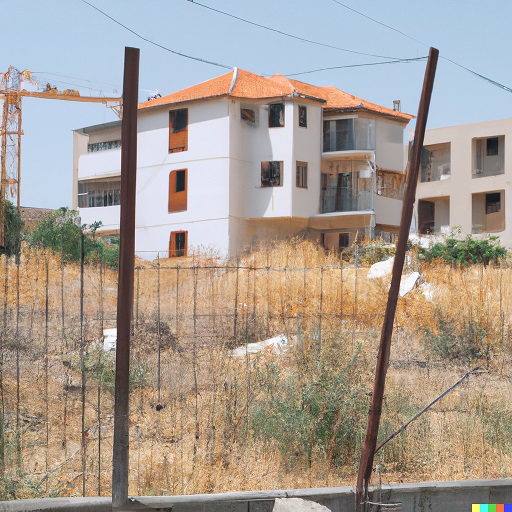} \\
            (c) & (d) & (g) & (h) \\
        \end{tabular}%
\end{table*}

The ability of the model to capture potential causal effects is well portrayed in table~\ref{tab:pollution}(a,b) where pollution is clearly associated with an active industrial area clearly generating copious amounts of smoke. In contrast, table~\ref{tab:pollution}(d-f) portray residential suburban areas with some green space, which are naturally associated with higher air quality.
\begin{table}
    \centering
    \caption{High air pollution (Top) vs clean air (bottom)\label{tab:pollution}}
        \begin{tabular}{ccc}
             \includegraphics[width=0.3\linewidth]{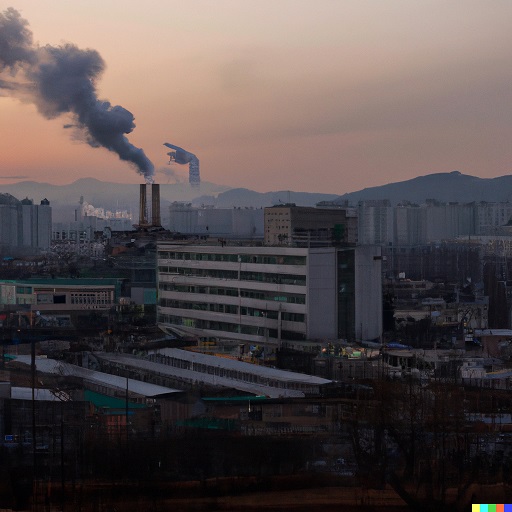}
            & \includegraphics[width=0.3\linewidth]{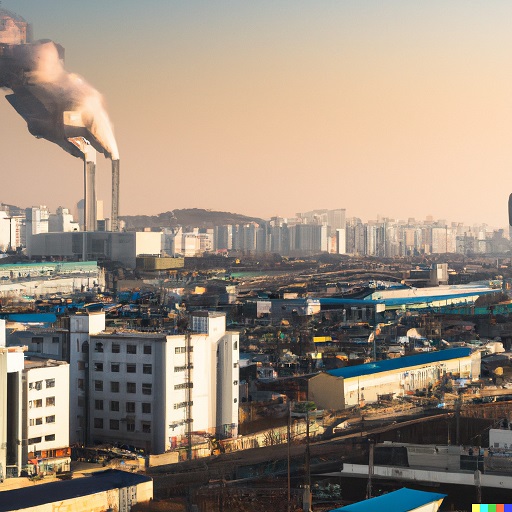}
            & \includegraphics[width=0.3\linewidth]{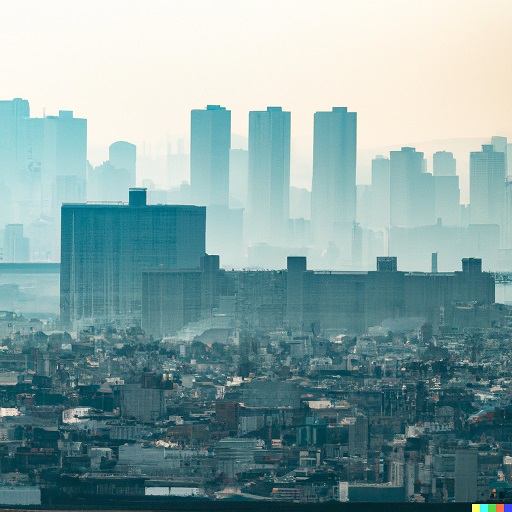}\\
            (a) & (b) & (c) \\
             \includegraphics[width=0.3\linewidth]{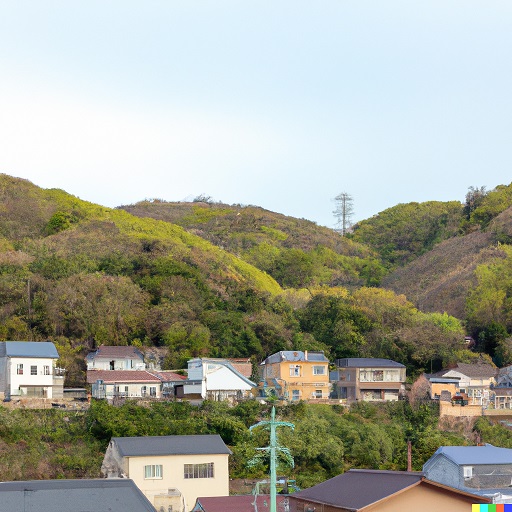}
            & \includegraphics[width=0.3\linewidth]{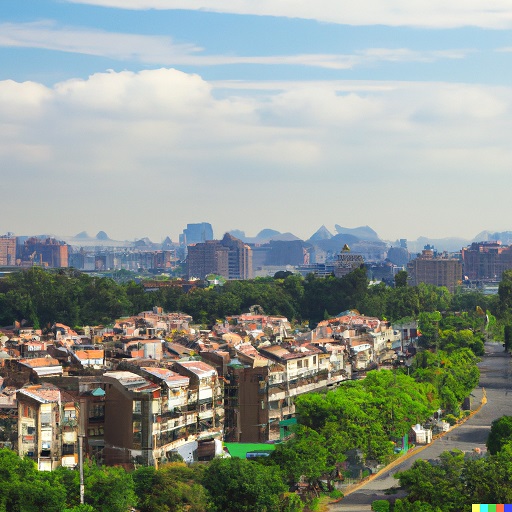}
            & \includegraphics[width=0.3\linewidth]{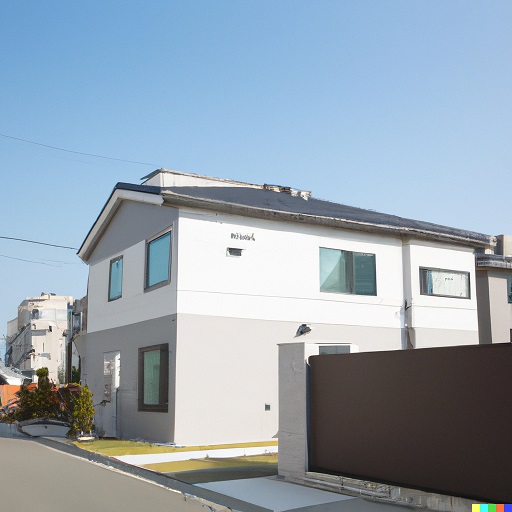}\\
            (d) & (e) & (f) \\[-2pt]
        \end{tabular}%
\end{table}

Futher observations are as follows:
\begin{itemize}
    \item The prompt `A painting of G' often captures complex relationships in a more interpretable form and more consistently than realistic scenes of most built environments, where G forms the grammar for a natural built environment scene. We hypothesize that this is because the painting domain inherently contains more variation (and is thus more forgiving) of variation within the image than real scenes (which tend to be more coherent and consistent). Prompts like `a cartoon of G' or `a poster of G' also produced coherent content, with the complexity of the content following the descending order of painting, cartoon, poster.
    \item Text generated within an image appeared to be nonsensical, but consistent. For example, the word `neighborhood' as a prompt was often associated with terms such as `neyehrood' or `neothrood' appearing as text in the generated image. This nonsensical text appears to refer to the word 'neighborhood' and such effects have been observed elsewhere~\cite{daras2022discovering}. In our work, we ignore all text generated within an image.
    \item Specialist terms from the built environment literature are often poorly understood by the model. Examples include terms such as `walkability' or `walkable' as an adjective. This can be attributed to the rarity of such terms appearing as captions in standard datasets.
    \item Prompts with the potential to generate disturbing imagery were completely avoided, where such imagery was unintentionally generated they were immediately filtered from the collected dataset along with the prompt which produced them. This included terms related to fire, overt references to violence, and generally any prompts which could lead to content unsuitable or irrelevant for analysis of the built environment.

\end{itemize}

We additionally generated 11000 images using Stable Diffusion\cite{Rombach_2022_CVPR}. While these were not analysed using the in-painting workflow nor qualitatively by inspection, they are included as part of the text-image pair data-set released as part of this work.

\section{Discussion and Future work}

We find that the different format domains explored (natural, painting, cartoon, poster, model, sketchup) capture different levels of detail and have differing abilities to represent the complexity of natural built environment scenes. This may generalize to other domains or disciplines as well.

The ability to capture realistic shadows and behavior of light was well demonstrated (primarily observed in natural imagery and the painting domain - example in Fig~\ref{fig:painting_walking}). This has implications for capturing shade requirements for aspects such as thermal comfort in open spaces in urban design (especially with further finetuning of the model). 

The images generated from the available DALL-E 2 framework are built on sampling from a latent space that is trained on a highly diverse set of images and captions, and architectural or urban design represents an extremely specialised niche of the images that can be generated through DALL-E 2. Although this is a considerably large model (3.5Bn Parameters), an inevitable trade-off between the diversity of contexts and the amount of information that can be captured from a given context may exert a limit on the level of details that DALL-E 2 captures. A future study may consider expanding the ability of DALL-E 2 to train specifically on a dataset with high-level of architectural details, and more nuanced and contextual captions. A more novel approach may investigate modifying the classifier-free guidance mechanisms to be conditioned on the context, in addition to the text captions. 

Many biases were observed in the scene generation process, particularly for natural imagery scenes. There is a bias towards generating western, urban scenes with blue skies and sunny weather.

While we ignored all text generated within speech bubbles in the cartoon domain, future work may be able to map these into more interpretable forms as first attempted by~\cite{daras2022discovering}. This may provide a more robust understanding of the intent of the image. Appending the prompt "with subtitles" to the grammar will also be helpful for such analysis.

The method is able to capture generalization to different geographic locales with remarkable realism. However, certain conditions associated with locales may not be appropriately represented. For example, a prompt for natural scenes ending with 'Aomori city, Japan' (known to be a snowy city) is not represented with snow, whereas a prompt for Antarctica is. For comparison, a Google search for ``Aomori city, Japan" retrieves primarily images with snow while a prompt for only ``Aomori city, Japan" in the model generates some images with snow. Overall the model shows a bias towards sunny scenes. Due to the model's robust ability to generalize for many domains (including weather), however, it is trivial to fix such `mistakes' using additional prompts such as "with snow" or ``in winter".

DALL-E 2 can capture the human perception of various aspects of the built environment. We observe that many positive urban qualities (such as `healthy’, `safe’, `happy’) generate similar images, notably with abundant greenery. These observations follow the surveyed human perception of such neighbourhoods as suggested in \cite{Herrmann-Lunecke2021} and \cite{Pfeiffer2016}. Results from DALL-E 2 can be said to reflect the human perception because, ultimately, the model is trained on human-generated knowledge. We raise the possibility for DALL-E 2 and the like to be a tool for gathering the human conceptualisation of certain aspects that designers would otherwise test through population surveys.

\section{Conclusion}

In this work, we evaluate the potential for transformer-based text-to-image models to design realistic images and scenes relevant to built environment disciplines such as urban planning. We find that the evaluated model DALL-E 2 has significant potential to aid human experts in this area, without performing any additional finetuning on a domain/task-specific dataset. We find that the evaluated model can effectively generate images in multiple domains relevant to urban scenes, including natural real-world scenes, abstract representations such as posters, paintings, and cartoons as well as realistic renders in relevant format domains such as sketchup, architectural models, and minecraft. We find that the model's weaknesses include creating realistic real-world scenes with a high level of detail. The model also struggles with compositionality in some scenes. Text prompts featuring forms of negation can also fail. Overall, our assessment reveals that text-to-image models may significantly speed up the design process of certain disciplines (such as graphic design and artistry), perhaps even rendering some existing processes obsolete. For more evidence-based and scientific design disciplines such as urban planning and design, our assessment indicates that further domain-specific training may be required. Even so, the model is able to capture a startling amount of existing knowledge from both the literature and best practices in the areas of urban planning and design. We release our dataset to encourage broader cross-disciplinary discussion in the space of automatic image generation for design disciplines: \href{https://github.com/sachith500/DALLEURBAN}{https://github.com/sachith500/DALLEURBAN}.

\section*{Acknowledgments}

This project is supported by National Health and Medical Research Council Grant GA80134. We would like to thank the anonymous reviewers for their insightful feedback. 

{ \small
\bibliography{references}
\bibliographystyle{plain}
}

\end{document}